\documentclass[11pt,letterpaper]{article}

\usepackage[hyperref]{acl2019}
\usepackage{url}
\usepackage{times}
\usepackage{latexsym}
\usepackage{graphicx}
\usepackage{longtable}
\usepackage{tabularx}
\usepackage{xtab,booktabs}
\usepackage{caption}
\aclfinalcopy % Uncomment this line for the final submission

\title{The Mafiascum Dataset: A Large Text Corpus for Deception Detection}
\author{Bob de Ruiter \\ Department of Computer Science \\  Radboud University \\ Nijmegen, the Netherlands \\ {\tt bob.de.ruiter@student.ru.nl}
         \And  
         George Kachergis \\ Department of Psychology \\ 
         Stanford University  \\ 
         Palo Alto, CA \\
         {\tt gkacherg@stanford.edu}}

\date{}

\begin{document}

\maketitle

\begin{abstract}
Detecting deception in natural language has a wide variety of applications, but because of its hidden nature there are currently no public, large-scale sources of labeled deceptive text.
This work introduces the Mafiascum dataset\footnote{\url{https://bitbucket.org/bopjesvla/thesis/src}}, a collection of over 700 games of Mafia, in which players are randomly assigned either deceptive or non-deceptive roles and then interact via forum postings. Over 9000 documents were compiled from the dataset, which each contained all messages written by a single player in a single game.
This corpus was used to construct a set of hand-picked linguistic features based on prior deception research, as well as a set of average word vectors enriched with subword information.
A logistic regression classifier fit on a combination of these feature sets achieved an average precision of 0.39 (chance = 0.26) and an AUROC of 0.68 on 5000+ word documents. On 50+ word documents, an average precision of 0.29 (chance = 0.23) and an AUROC of 0.59 was achieved.
% 147 words
\end{abstract}

\section{Introduction}

A reliable automatic deception detector for written communication would find wide application in intelligence agencies, law enforcement, and online marketplaces.
Of course, deception is hard to find and label because of its deceitful nature: deceivers are aiming to not be caught, and as such do not advertise their successes. Moreover, it is often unclear whether falsehoods are deliberate, and, in the case of spam, whether deception is deliberately transparent \cite{herley2012nigerian}.
As such, a major hurdle in applying supervised learning techniques to deception detection is the lack of a suitable publicly available dataset.

Although deception takes place in a large number of datasets, such as court transcripts \cite{perez2015deception}, the Enron email corpus \cite{keila2005detecting}, and laboratory experiments \cite{perez2015experiments}, to date all public labeled datasets comparable in size to high-profile sentiment analysis benchmarks \cite[e.g.][]{rosenthal2017semeval} focus on single-sentence non-interactive deception \cite[e.g.][]{perez2015deception, wang2017liar}.
In this paper, we introduce a naturalistic dataset, a collection of over 700 games of Mafia played on an Internet forum, in which players are assigned either a deceptive or a non-deceptive role. The final dataset consists of over 9000 documents, which each contains all messages written by a single player in a single game. The average document contains 3940 words.
We test a variety of established linguistic cues to deception on this dataset, as well as testing word embeddings.

\subsection{Linguistic Cues to Deception}\label{deception-research}
% Research shows that people are not very good at discriminating between truths and lies (C. F. Bond & DePaulo, 2006)
Until the 1970s, research aimed at detecting deception was largely focused on finding nonverbal cues (e.g., body language and facial expressions) for use in face-to-face interactions. 
After \citet{Mehrabian:1971} found that slow, erroneous and sparse speech can indicate deception, systemic research into linguistic cues to deception took off. 

In a recent meta-analysis of 79 linguistic cues to deception from 44 studies, \citet{hauch2015computers} found that deceivers express more negative emotions, distance themselves from events (using less perceptual and sensory language), and generally appear to experience more cognitive load as compared to non-deceivers.
More specifically, deceivers produced fewer words and fewer distinct words than truth-tellers, while using more sentences.
Additionally, deception was found to correlate with the use of negation terms (e.g., `no', `never', and `not') and words specifically related to negative emotions and anger (e.g., `hate', `worthless', `enemy').
Deceivers used fewer exclusive words (e.g., `but', `or', and `without') than non-deceivers, as well as fewer tentative words (e.g., `may', `seem', and `perhaps').
Deceivers used fewer total first-person pronouns than truth-tellers, but used second- and third-person pronouns more often than truth-tellers do.
Deceivers also used fewer sensory and perceptual details than truth-tellers, especially in the acoustic realm (e.g., `listen', `sound', or `speak'), as opposed to sight or feeling. 
Deceivers produced more motion verbs (e.g., walk, go, or move).
Finally, compared to truth-tellers deceivers also used fewer words concerning their inner thoughts (insight) and cognitive processes.

The meta-analysis supported the observation by \citet{hancock2013eye} that the significance of linguistic cues to deception is heavily dependent on context. In particular, effect sizes varied wildly across different interaction conditions, types of deception, modes of communication, deceivers' motivations, and differences in emotional valence. In one case, even the direction of the effect of a significant linguistic cue differed between contexts. Studies included in the meta-analysis skew towards contexts that are different from online Mafia in terms of setting, mode of communication, interactivity, deception type, and possibly participant demographics, limiting direct comparison to our study, but it is the only comprehensive survey of systematic research of cues to deception we know of.

% truth-tellers: word quantity; exclusive words; tentative words; 
%words referring to sensory-perceptual processes, cognitive processes and insight; the ratio of unique words over all words, also known as the type-token ratio; quantifiers; and first-person pronouns.
%deceivers: sentence quantity, negations, negative emotion words, anger, and second-person and third-person pronouns

% Miscellaneous Category (p. 324) subsection also contains several significant cues...

\subsection{The Game of Mafia}\label{the-game-of-mafia}

The original face-to-face version of the game of Mafia, also known as Werewolf, was designed by Dmitry Davidoff in the 1980s to model a conspiracy of an informed minority, the Mafia, within an uninformed majority, the townsfolk. 
Before the game starts, every player is randomly assigned to one of these teams.
The goal of the Mafia, who, unlike the townsfolk, are aware each others' identities, is to vote out all townsfolk, while the town's objective is to vote out all Mafiosi. 
Being eliminated precludes townsfolk nor Mafia from winning with their team, to prevent conflict of interest.

The game starts with an in-game Day, during which the players discuss their suspicions, or, in the case of the Mafia, pretend to, and try to agree on a vote. 
When a majority vote is reached on a player, that
player is eliminated from the game, their role and alignment are
announced publicly by the moderator, and the game goes on to the Night.
In games without special roles, also known as vanilla or mountainous, the only action performed during the Night is the elimination of another player by the Mafia. 
In face-to-face games, this is usually done by the
Mafia pointing at the target while the townsfolk have their eyes closed.
When the Mafia have reached a non-verbal agreement and their eyes are closed again, the game moderator announces the name, role, and alignment of the eliminated player, everyone ``wakes up'' and the game goes on to the next Day.

Mafia has been adopted by a large number of online communities. 
On the Internet, Days are usually played out in a
public forum thread or chat room and the Mafia decide on the eliminated player by private messages.

An experiment run by \citet{Zhou:2004} shares multiple similarities with later studies that use the online game of Mafia as a model for deception, which includes our study. 
Rather than Mafia, another consensus-building task was used, in which the goal is to persuade another person of a plan in a hypothetical survival scenario. 
The test group was told to convince their partner of a predetermined solution they were told was incorrect, while the control group was instructed to argue for their true views. 
Task communication occurred outside the lab over email, across a timespan of several days. 
The authors noted that this reduced the amount of experimental control, but also in their eyes lessened the pressure and unnaturalness felt by participants, perhaps making the experiment more representative of real-life deception. 
Participation was compulsory for students, and no additional reward was given for successful deception.

In contrast with prior research, the study found that deceivers wrote more than truth-tellers. 
The authors pointed to the persuasive nature of the task, which requires deceivers to come up with arguments to support their claims. 
Another possible explanation, not discussed by the authors, is that the task was more interesting for deceivers than for the control group, causing a difference in the amount of effort put in.

\citet{Zhou:2008} collected 1192 Mafia games from a popular Chinese website dedicated to this game. 
The dataset differs from the Mafiascum
dataset used in our study, which is explored in detail below, in many respects.
All players were Chinese and all messages were written in Chinese. 
The deadline to decide on a single elimination was 3 minutes, during which all players were expected to stay in the chat room. 
The dataset only included games with a size ranging from 6 to 8 players, of which only one player was a deceiver.

In this setting, the deceivers' average word count was found to be low compared to truth-tellers, which is inconsistent with \citet{Zhou:2004}, but consistent with most other previous research. 
Additionally, the vocabulary of the deceivers tended to be more diverse, which is inconsistent with both \citet{Zhou:2004} and other previous research.
The authors attributed these inconsistencies to cultural differences between the Chinese Mafia players and American students and differences between email and chat rooms, but they could also be the result of higher engagement from the truth-tellers: since the Chinese players chose to play voluntarily, it is reasonable to assume that they were
more invested in hunting for deceptive players than the students who were required to participate in an experiment and who did not even know they might be deceived.

\subsection{Machine Learning Classification}\label{machine-learning-classification}

The performance of many recent text classification techniques on deception datasets is unknown. This is unfortunate for those interested in systematically detecting deception, but also for those in the business of creating general text classification techniques. Intuitively, deception detection is quite different from most text classification problems, as it does not allow classifiers to base their predictions solely on explicit information the author intended to convey, such as their opinion on a movie. Instead, it requires classifiers to find implicit information the author intended to hide.

In the case of Mafia, the deceiving condition is randomly assigned to players in the game of Mafia, whereas many popular text classification benchmarks, such as most large sentiment analysis datasets, are passively observed. 
Because of this, the negative signal a performant sentiment classifier picks up on may not be negative sentiment, but the writing style of the type of person who publishes
negative movie reviews on the Internet, for example.

Much of the less recent machine learning research on deception used Support Vector Machines. 
\citet{Mihalcea:2009} collected data
from three written deception tasks. 
Applying only basic stemming and using nothing but raw stem counts, a Support Vector Machine
trained on one of the tasks correctly classified 70\% of the documents from that task on average. 
Using the same setup, a Naive Bayes classifier reached 71\%. 
Additionally, an SVM trained on two tasks was able to correctly classify 58\% of the documents from the third task, while the Naive Bayes classifier reached a classification rate of 60\%.

State-of-the-art document classification techniques that have not been used in deception detection tasks include deep neural networks and word embeddings enriched with subword information \cite{Bojanowski:2016}. 
The latter is promising because of its performance on syntactic tasks, since many cues to deception are plain syntactic groups. 
The FastText project provides word embeddings enriched with subword information for 294 languages, opening up the possibility to transfer some of the methods in this paper to deception in other languages.
Regular word embeddings \citep{Bengio:2003} have been used to detect deception, producing results comparable to simpler techniques \cite{Mihaylov:2016}, although this may be explained by the fact that the dataset used in the relevant study was not randomized with large differences between groups.
% \citet{Ren:2016} compare a simple neural document model using paragraph vectors and Gated Neural Networks (GNNs), and finds that GNNs outperform paragraph vectors, which is consistent with the result found in sentiment classification \cite{Tang:2015}.

In this paper, hand-picked text features that have previously been proven to be successful in classifying deception are compared to average word vectors enriched with subword information.
Additionally, we define a benchmark for text classification pipelines on the Mafiascum dataset.

\section{The Dataset}

%In addition to providing the information required to understand these caveats, we briefly review the other notable online Mafia archives to explain why we chose to scrape Normal games from Mafiascum.

%The first known game of online Mafia was played on Internet forum The \href{http://www.greylabyrinth.com/discussion/viewforum.php?f=15}{Grey Labyrinth} in 2000. 
%The Days were played out in a public forum thread and the Mafia decided on the eliminated player by private messages.
%Games on this forum are suitable for deception research, but small in number.

In 2002, Mafiascum\footnote{\url{https://forum.mafiascum.net}}, a forum dedicated to games of Mafia and discussion of Mafia theory, was started. It has six million posts. In-game Day phases on Mafiascum usually last two weeks, during which players are expected to post at least every 48 hours. 
If a player cannot play anymore because of unforeseen
circumstances, their slot is filled by a new player who is expected to read the entire game before continuing in their place.

Although Mafiascum currently runs four types of Mafia games, only the Normal archives include easily parsable alignment distributions for almost all games. Normal games are characterized by the use of a limited set of well-known roles and mechanics. This is fortunate, since extreme deviations to Mafia present in some non-Normal games may introduce linguistic noise that is unrelated to deception.

Still, a few noise-introducing deviations are present in our dataset. Until 2014, multiple players were allowed to play as a single player under a single account. Game moderators have always been allowed to put multiple competing minority factions in one game.

Although Mafiascum remains active to date, real-time games of Mafia have surpassed forum games in popularity.
Unfortunately, most real-time games of Mafia have a high number of additional roles, making the game less suitable as a linguistic model for deception. 
For example, on Epicmafia\footnote{\url{https://epicmafia.com}}, which has been used in non-linguistic deception research \cite{Pak:2012}, the optimal strategy for players with special roles is often to immediately publicly claim to have such a role, confirming their alignment. 
In other games, the optimal strategy for the Mafia is to have one Mafia member falsely claim to have a leading role themselves, effectively confirming to the townsfolk that one of two players is Mafia. In most Normal games on Mafiascum, the list of roles and alignments present in the game is not known to players.

Critically, in these games, deceptive language is not the only signal a classifier can detect to differentiate between the Mafia and the townsfolk. 
A classifier that can detect uncontested role claims could
correctly identify some townsfolk, not based on their linguistic signature, but based on the fact that the Mafia cannot make uncontested role claims in most Epicmafia games. 
In the Mafiascum dataset, the extent of this problem is smaller, since absolute public confirmation of non-eliminated players is relatively rare. Exact numbers are hard to come by, but in a random sample of five thirteen-player games, no such confirmations occurred.

\subsection{Investigation of Possible Confounders}

In some online gaming communities, a small number of unusually active players account for a disproportionately large part of produced data, which may bias a classifier to perform well only on this small set of players. 
After preprocessing, our dataset includes 9676 documents from 685 games. 
The most active user account has played in 57 games, accounting for an equal number of documents in the dataset. 
It should be noted that many active players
have less-used alternative accounts, making this a lower bound.
Nonetheless, looking at the activity distribution, we expect that no player has played in more than 20\% of all games, accounting for no more than 1.6\% of all documents.

%\begin{figure}
%\centering
%\includegraphics{../thesis/games_played_hist.png}
%\caption{The vast majority of users has played less than 10 games}\label{games}
%\end{figure}

\begin{figure}
\centering
\includegraphics[width=0.5\textwidth]{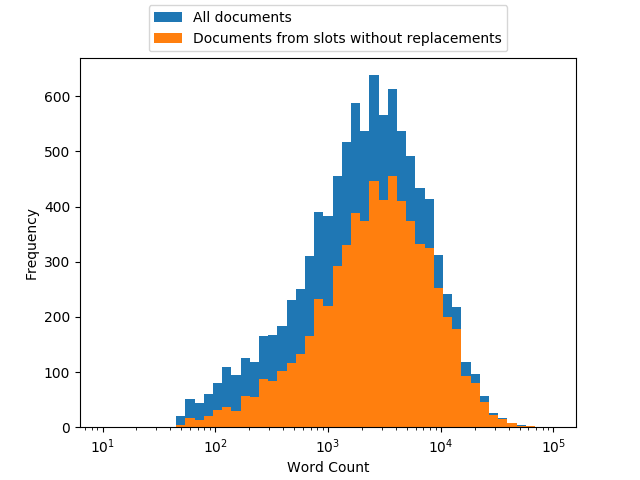}
\caption{Histogram of word counts per user per game}\label{wc_hist}
\end{figure}

Another concern pertains to the fact that Mafia are slightly more likely
to be replaced than townsfolk. 
In our dataset, an average town-aligned slot has 0.33 replacements, while an average Mafia-aligned slot has 0.35
replacements, meaning that a replacement is 6\% more likely to be Mafia. 
Since replacements often use very distinctive language, providing comments on the entirety of the game when they first catch up, we expect that replacement detection can reliably be used as a proxy for deception. 
Throwing out all documents from slots with replacements
would solve this, but it is a harsh measure, since they make up a sizable portion of the dataset, even among the larger documents (\autoref{wc_hist}). In the results section, we show that the removal of replacement documents does not decrease the overall performance, at least not for our simple linear model. Because of this, replacement documents were included in the published dataset.

Although player alignments are randomized within games, the
townsfolk-Mafia ratio is decided by moderators on a per-game basis. As the game
progresses, it may become clear to players that a large number of Mafia
are in the game, either because of setup speculation or because players
of multiple Mafia factions have been eliminated already. Player may
discuss this: ``This game probably has lots of Mafia.'' If such a
phrase were to be used consistently by all players in games with a high
Mafia-townsfolk ratio, including the Mafia, a classifier would assume
this phrase to be indicative of deception. This is not necessarily
a confounder: in real life, deception occurs more in some groups than
others, and people's speculation on this matter may help deception
detection. However, it may hurt the generalizability of the classifier.

According to veteran players, the way Mafia is played on Mafiascum has
shifted over time. This includes high-level subjective criteria such as
the preferred way to start discussion at the start of the game, popular
modes of analysis, and the extent to which players express emotion, but
it also includes the average post count and post length in a game. While
communication was very structured and reminiscent of debating in the
early days, bursts of stream-of-consciousness posting have become more
and more common over time. Initially, we expected the townsfolk-Mafia ratio to have
varied over time as well, since setup design has changed in many ways
since 2002. Fortunately, the townsfolk-Mafia ratio has remained
consistent over time, making time period an unlikely confounding variable.

Another possible confounder that ended up being moot was game size.
Large games typically have more experienced players, more replacements,
and a higher game length than small games, all of which could be
reflected in language use. We also expected game size to influence the
dependent variable, since the percentage of players aligned to the Mafia
needed to balance a vanilla game of Mafia decreases as game size
increases \cite{Migdal:2010}. Surprisingly, this effect seems to be entirely
counteracted by the special roles and mechanics on Mafiascum; no
relation between the percentage of Mafia members and game size exists.

Finally, one should keep in mind that although there is no known public
record of demographics, the Mafiascum userbase is unlikely to be
representative of the global population or the population of regular
participants of scientific studies.

\subsection{Unpublished Work by Mafiascum Members}

Multiple Mafiascum members have performed systemic analyses of cues to
deception. In a private forum, Mafiascum user Psyche posted that logistic
regression, multinomial naive Bayes classification, binomial naive
Bayes, and an ensemble performed no better than random on a recursive
feature selection of an unspecified feature set. Equal numbers of town and
scum were taken from each game.

Mafiascum user \citet{goodmorning}
found that Mafia are just as likely as townsfolk to be part of an
elimination vote on another Mafia member in games for new players. Also
in games for new players, Mafiascum user
\citet{toomai}
found that the Mafia elimination rate is higher than chance in the
majority of game states, although there are states, such as the start of
the game, where random and directed lynching have equal performance.
Even in game states where the elimination rate is better than chance,
this is not necessarily caused by good player judgment, as it could
also be the result of the special roles included in games for new
players.

Mafiascum user \citet{loopdan} found that new players are significantly more likely to
be aligned to the Mafia if they post a greeting to all players at the start of the game. An attempted replication on our dataset, which was mostly produced by non-new players, failed. The occurrence of generic greetings appeared to be much lower in our dataset. This may be the result of experienced players being more composed, or more aware of how games typically start.

\section{Machine Learning Benchmark}

The Mafiascum dataset seems promising not only for deception research but also as a general benchmark for supervised text classification pipelines. We propose the following benchmark task: Take a twenty-fold stratified shuffle split of all documents with a word count of 50 or higher. For every fold, fit a fresh instance of the pipeline on the training set. Use that instance to generate predictions for the test set. These predictions should at least be scored using the area under the precision-recall curve. The baseline score set by this paper is 0.286. A Python implementation of this benchmark is included with the dataset.

\section{Methods}\label{methods}
Games of Mafia and alignment distributions were scraped from Mafiascum's Normal Game archives. 
The scraping code, which uses the JavaScript
bookmarklet artoo.js, and the resulting JSON output are included in the dataset\footnote{\url{https://bitbucket.org/bopjesvla/thesis/src}} along with the output of the preprocessing and feature construction steps described below.

\subsection{Preprocessing}\label{preprocessing}

Games that did not have a complete alignment distribution available were discarded.

Since games are moderated by humans, there is a period between a conclusive elimination vote and the role reveal, which is known as twilight. 
Eliminated townsfolk usually speak freely during this period, but eliminated Mafia members often clam up, wary of giving the town additional information on their partners. The exception to this is the moments after the final elimination, when many players have often
revealed their alignment before the moderator has officially declared a win. 
In any case, seeing that eliminated players have no incentive to keep up their personal facade, post-elimination and post-game posts should not be included in the dataset. 
As such, all posts after the final vote count, which signifies the end of the game, were discarded.
Additionally, to catch posts from eliminated players in
twilight, all posts a player made in the 24 hours before their last in-game post, including their last in-game post, was also discarded. A 24-hour cut-off was chosen because moderators typically check in on a game at least once a day.

For every player in every game, the remaining in-game posts were merged into a single text document. 
If the remaining number of words in a document was lower than 50, the document was discarded. 
Each document was assigned a binary label, signifying whether or not the document's author was Mafia.
Documents from players with a role named the Serial Killer, a third-party lone wolf faction, were removed.

If multiple players occupied a slot at different times because of player replacement, their posts were put in separate documents, but if two players played together under a single account, this could not
automatically be detected. 
As a result, a minority of the documents contain posts written by different players.
This introduces no systemic bias, but it may introduce noise.

\subsection{Feature Construction}\label{features}

All remaining documents were split into words, where a word is defined as a series of characters marked as letters in the Unicode database, which includes numbers.
% punctuation removed? Yeah, cleared this up.
Most hand-picked features were adapted from the meta-analysis by \citet{hauch2015computers}, with the exception of two features specific to the communication medium: message length and the number of messages sent per 24 hours. Word category frequencies were computed for word category cues the meta-analysis reported significant effects on, which are described in the introduction.

Instead of the frequency of exclusive words, however, we computed the term frequencies for the words ``but'' and ``or''. Because exclusive words are less related to each other than the other word categories that we use to identify deception, such as first-person pronouns, we believe distinct exclusive words may relate to deception in different ways. 
The counts of other exclusive words were not used as features, since ``but'' and ``or'' are much more common than all other exclusive words combined.  Therefore, the signal found in previous research likely originated from one of these subfeatures.

The word count per 24 hours, the average token length, and the average sentence length were also computed for every document. Every hand-picked feature was scaled based on its variance. An overview of all features along with their correlations to truth-telling is shown in \autoref{r}.
%TODO: relate this to meta-analysis. Done!

\begin{table}[ht!]
\begin{tabular}{lr}
\toprule
Feature & AUROC\tabularnewline
\midrule

\textbf{Message length} & \textbf{0.552} \tabularnewline
\textbf{Messages / 24 hrs} & \textbf{0.453} \tabularnewline % msg24h ?
\midrule
Word Count / 24 hrs & 0.490  \tabularnewline
``or'' ratio & 0.501 \tabularnewline
\textbf{3rd-person pronoun ratio} & \textbf{0.533} \tabularnewline
\textbf{2nd-person pronoun ratio} & \textbf{0.476} \tabularnewline
First-person pronoun ratio & 0.498 \tabularnewline
\textbf{``but'' ratio} & \textbf{0.543} \tabularnewline
Token length & 0.506 \tabularnewline
Type-token ratio & 0.487 \tabularnewline
\textbf{Sentence length} & \textbf{0.453} \tabularnewline
Anger ratio & 0.496 \tabularnewline % angry words? or sentiment?
\textbf{Sensory words ratio} & \textbf{0.520} \tabularnewline
Cognitive process words ratio & 0.503 \tabularnewline
Insight word ratio & 0.506 \tabularnewline % what's this?
Motion word ratio & 0.491 \tabularnewline % list of motion words?
``not'' ratio & 0.497 \tabularnewline
\textbf{Quantifiers ratio} & \textbf{0.532} \tabularnewline
Negative emotion words ratio & 0.499 \tabularnewline
Tentative words ratio & 0.513 \tabularnewline
\end{tabular}
\caption{AUROCs, or common language effect sizes, for linguistic variables. 
Values smaller than 0.5 indicate higher prevalence among deceivers. 
Boldface indicates a significant Mann-Whitney U test. Features above the ruler are specific to forum communication. Features below the ruler are adapted from \citet{hauch2015computers}. Ratios are word (category) frequencies.}
\tabularnewline
% unique tokens is a weird term for Type/Token Ratio (TTR)
\label{r}
\end{table}

Pretrained 300-dimensional word vectors with subword information, trained on the English Wikipedia, were obtained from \citet{ft}. The words in each document were mapped to their corresponding word vectors, using the subword information to compute word vectors for out-of-vocabulary words. After this, we computed the average of all word vectors for each document, resulting in a single vector with a dimensionality of 300 per document.

A third feature set was created by repeating the previous procedure, this time using word vectors without subword information trained on Wikipedia and Gigaword, obtained from \citet{Pennington2015Nov}. In this case, out-of-vocabulary words had to be discarded.

Two more feature sets were created by concatenating the hand-picked features to each of the word vector feature sets.

\subsection{Machine Learning Benchmark}\label{machine-learning-statistical-analysis}

Implementing the benchmark described above, a regularized (C = 1.0) logistic regression model was trained and tested on a stratified 20-fold split of each feature set. For training, the two classes were reweighted using the heuristic devised by \citet{king2001logistic}.

In addition to class weighting, we also reweighted individual training samples based on their word count. The intuition behind this is simple: a 10000-word deceptive document typically contains more information about deception than a 100-word document, but not as much as 100 independent 100-word documents. Because no literature could be found on this type of reweighting, we took a conservative approach: the sample weight of a document was set to the log of the word count, meaning that the weight of a 10000-word document was set twice as high as the weight of a 100-word document of the same class.

% hidden because of space constraints
Experiments with paragraph vectors and single-post classification (as opposed to the practice of concatenating multiple posts into documents) were abandoned in early stages as they did not seem to hold any promise.

\section{Results}

\subsection{Statistical Analysis}\label{statistical-analysis}

% Except for message length and the number of messages sent per 24 hours (which were respectively negatively and positively correlated with truthful roles), all features in \autoref{r} were either taken or adapted from well-performing cues in the primary meta-analysis in \citet{hauch2015computers}.
Of the 18 features adapted from the cues \citet{hauch2015computers} found to have significant effects, 6 were also significant in the Mafiascum dataset, although all effect sizes were small (\autoref{r}). Of those 6, only 2 features, sentence length and the ratio of third-person pronouns, had the same direction as in the primary meta-analysis. The unexpected result that four features replicated in the opposite direction is investigated in the discussion.

Of all features that were originally introduced in deception research under the assumption that complexity is a proxy for truthfulness, sentence length was the only one to have the expected negative correlation to deception.
Deception was not correlated with word length, nor with the type-token ratio. Neither of the exclusive words indicated truthfulness. Usage of the word ``but'' correlated positively with deception, while usage of the word ``or'' did not predict anything. This casts doubt on the common practice of grouping exclusive words together in linguistic deception research.

Deceptive roles were significantly positively correlated with post length, but negatively correlated with post frequency. 
No significant correlation between word count and deception was found.

\subsection{Machine Learning}

\begin{table}[ht!]
\begin{tabular}{lll}
\toprule
Feature set &  AUROC &     AP \\
\midrule
   Hand-picked &  0.566 [0.552, 0.579] &  0.270 \\
      FastText &  0.578 [0.565, 0.592] &  0.279 \\
 HP + FastText &  0.593 [0.579, 0.606] &  0.286 \\
         GloVe &  0.572 [0.558, 0.585] &  0.275 \\
    HP + GloVe &   0.583 [0.570, 0.597] &  0.285 \\
\midrule
 HP + FT - repl &  0.596 [0.579, 0.612] &  0.280 \\
\bottomrule
\end{tabular}
\caption{Cross-validated performance of the models trained on different feature sets (AUROC = area under the Receiver Operating Characteristic curve including 95\% confidence intervals, AP = average precision [chance = 0.23], ``- repl'' = replacements removed from train and test set)}\tabularnewline
\label{results}
\end{table}

\begin{table}[ht!]
\begin{tabular}{llrr}
\toprule
Word count &     N &  AUROC &     AP \\
\midrule
        50+ &  9676 &  0.593 &  0.286 \\
      5000+ &  2401 &  0.651 &  0.344 \\
\midrule
     50-999 &  2592 &  0.547 &  0.250 \\
  1000-2999 &  3120 &  0.589 &  0.292 \\
  3000-4999 &  1563 &  0.603 &  0.308 \\
  5000-6999 &   824 &  0.678 &  0.385 \\
  7000-8999 &   538 &  0.678 &  0.360 \\
 9000-10999 &   310 &  0.587 &  0.301 \\
     11000+ &   729 &  0.623 &  0.322 \\
\bottomrule
\end{tabular}
\caption{Cross-validated performance of the model trained on HP + FastText by word count segment}\tabularnewline
\label{range_res}
\end{table}

The logistic regression model performed significantly better than chance on all feature sets (\autoref{results}). No model trained on one feature set performed significantly better than a model trained on another.

Removing documents from slots with player replacements did not decrease classifier performance, warranting their inclusion in the dataset despite the possibility of a weak confounding effect, as explained in the Dataset section.

\section{Discussion}

The Mafiascum dataset seems promising not only for deception research but also as a text classification benchmark.
The most worrisome hypothetical confounding variables between
games did not appear to require any controlling, and deceptive roles are randomly assigned to players per game.
Combined with the fact that deceptive
players attempt to hide their label, this dataset tests for aspects of text classification that most benchmark datasets do not.

All effect sizes of the hand-picked features were small, which may be due to the slow pace of the game or the experience many players already have playing as a deceiver. The latter property has real-world relevance, since many deceivers of interest are repeat offenders. Despite the small effect sizes, a logistic regression model trained on the hand-picked features performed significantly better than chance.

Of the six significant hand-picked features taken or adapted from the meta-analysis by \citet{hauch2015computers}, only two, sentence length (+) and the ratio of third-person pronouns (-), matched the direction of the effect found in the meta-analysis. The ratio of second-person pronouns; the ratio of sensory words; the ratio of the word ``but'', which is a subfeature of the more commonly used ratio of exclusive words; and the ratio of quantifiers did not match the direction of the effect found in the meta-analysis.

However, the authors of the meta-analysis notes that the link between quantifiers and truthfulness was based on only four studies. They urge the reader not to draw strong conclusions about this cue, considering that the theoretical foundation is off: all other cues based on descriptive words, such as prepositions, did not predict truth-telling.

Other discrepancies in this study might be the result of differences in context and willingness to deceive. In our dataset, second-person pronouns are used less by deceivers and the word "but" is used more, which might point at deceivers trying to persuade people, instead of pretending to communicate with them. This is also supported by the positive relation between deception and third-person pronouns. The reason for the correlation between sensory words and deception remains unclear.

% TODO: mention these explicitly by name, and the four that don't match: do all the summarization for the lazy reader. discussion should repeat and contextualize your most significant points.

%In future work, we plan to further examine these word vector models to uncover additional properties of deceptive language.

Two features specific to the communication medium were included: message length and message frequency. Deceptive roles were significantly positively correlated with message length, but negatively correlated with message frequency. 
Considering that no significant relation between word count per 24 hours and deception was found, this likely means that deceivers refrain from posting in certain situations, perhaps because they would prefer to hear a genuine opinion first, perhaps because they think any contribution they make is going to attract unwanted attention, or perhaps because they dislike playing as a deceiver.

In a similar vein, the shorter posts from townsfolk may be the result of relatively unfiltered expression. 
Townsfolk may believe that they can broadcast any idea they come up with the moment they come up with it, since they know their thoughts are genuine. 
Mafia may feel the need to add more detail to their personal narrative. 

% The theory that deception draws on the same resources as language use is supported by the negative correlation between deception and sentence length, but undermined by the larger positive correlation between deception and usage of the word ``but''. 
% In deception theory, making distinctions is a marker of cognitive complexity, and exclusive words are considered a marker of distinctions. 
% Thus, a tentative explanation for our results is that because the Mafia's interests are the opposite of the other players, they must go against the grain to be successful. 
% This may explain why no significant correlations were
% found for the other exclusive word, ``or'', but it does not make sense if the Mafia consider blending in to be their primary objective.

The results in this paper are yet another strike against the idea that a linear model fit on traditional linguistic markers of deception can generalize across deception contexts. There is a slim chance that a linear model trained on average word vectors does generalize across contexts, but in our mind, it seems more likely that a model capable of capturing the interactions between deception context and word usage is required. % any relevant citations?

We expect that to train any model to detect deception across contexts, a large amount of deceptive and non-deceptive documents gathered from a multitude of contexts is needed. As such, we urge deception researchers to publish their datasets whenever possible.

% In future research on deception, word embedding classifiers trained on our dataset could be tested on existing deception datasets. 
% need a concluding paragraph -- I like this sentence or something like it as final, but the rest of the paragraph should restate progress, and discuss next steps (e.g., suggesting testing on multiple task types, as you had)
% Although a reliable, general linguistic deception detector would likely change the world in a profound way, a model that does not extend beyond relatively similar scenarios, such as computer-mediated interrogation settings, could also be of great use in branches of business and law enforcement.
% In future work, data from experiments on the other end of the deception spectrum, such as opinion falsification and imitation, but also data from experiments that are closely related to the game of Mafia, such as the persuasive task used by \citet{Zhou:2004}.

\bibliography{emnlp2017}

\begin{thebibliography}{}
\expandafter\ifx\csname natexlab\endcsname\relax\def\natexlab#1{#1}\fi

\bibitem[{Bengio et~al.(2003)Bengio, Ducharme, Vincent, and
  Jauvin}]{Bengio:2003}
Yoshua Bengio, Rjean Ducharme, Pascal Vincent, and Christian Jauvin. 2003.
\newblock A neural probabilistic language model.
\newblock {\em Journal of Machine Learning Research\/} 3(2):1137--1155.

\bibitem[{Bojanowski et~al.(2016)Bojanowski, Grave, Joulin, and
  Mikolov}]{Bojanowski:2016}
Piotr Bojanowski, Edouard Grave, Armand Joulin, and Tomas Mikolov. 2016.
\newblock \href{http://arxiv.org/abs/1607.04606v2}{Enriching word vectors with
  subword information}
  \href{http://arxiv.org/abs/1607.04606v2}{http://arxiv.org/abs/1607.04606v2}.

\bibitem[{FastText(2018)}]{ft}
FastText. 2018.
\newblock \href{https://fasttext.cc/docs/en/crawl-vectors.html}{{Word vectors
  for 157 languages {$\cdot$} fastText}}.
\newblock [Online; accessed 14. Feb. 2019].
\newblock
  \href{https://fasttext.cc/docs/en/crawl-vectors.html}{https://fasttext.cc/docs/en/crawl-vectors.html}.

\bibitem[{GloVe(2014)}]{Pennington2015Nov}
GloVe. 2014.
\newblock \href{https://nlp.stanford.edu/projects/glove}{{GloVe: Global Vectors
  for Word Representation}}.
\newblock [Online; accessed 14. Feb. 2019].
\newblock
  \href{https://nlp.stanford.edu/projects/glove}{https://nlp.stanford.edu/projects/glove}.

\bibitem[{goodmorning(2014)}]{goodmorning}
goodmorning. 2014.
\newblock \href{https://forum.mafiascum.net/viewtopic.php?f=5\&t=59076}{{VCA
  Stats {\ifmmode\bullet\else\textbullet\fi} Mafiascum.net}}.
\newblock [Online; accessed 5. Feb. 2019].
\newblock
  \href{https://forum.mafiascum.net/viewtopic.php?f=5\&t=59076}{https://forum.mafiascum.net/viewtopic.php?f=5\&t=59076}.

\bibitem[{Hancock and Woodworth(2013)}]{hancock2013eye}
Jeff Hancock and Michael Woodworth. 2013.
\newblock An “eye” for an “i”: The challenges and opportunities for
  spotting credibility in a digital world.
\newblock In {\em Applied issues in investigative interviewing, eyewitness
  memory, and credibility assessment\/}, Springer, pages 325--340.

\bibitem[{Hauch et~al.(2015)Hauch, Bland{\'o}n-Gitlin, Masip, and
  Sporer}]{hauch2015computers}
Valerie Hauch, Iris Bland{\'o}n-Gitlin, Jaume Masip, and Siegfried~L Sporer.
  2015.
\newblock Are computers effective lie detectors? a meta-analysis of linguistic
  cues to deception.
\newblock {\em Personality and Social Psychology Review\/} 19(4):307--342.

\bibitem[{Herley(2012)}]{herley2012nigerian}
Cormac Herley. 2012.
\newblock Why do nigerian scammers say they are from nigeria?
\newblock In {\em WEIS\/}.

\bibitem[{Keila and Skillicorn(2005)}]{keila2005detecting}
Parambir~S Keila and DB~Skillicorn. 2005.
\newblock Detecting unusual and deceptive communication in email.
\newblock In {\em Centers for Advanced Studies Conference\/}. pages 17--20.

\bibitem[{King and Zeng(2001)}]{king2001logistic}
Gary King and Langche Zeng. 2001.
\newblock Logistic regression in rare events data.
\newblock {\em Political analysis\/} 9(2):137--163.

\bibitem[{Loopdan(2019)}]{loopdan}
Loopdan. 2019.
\newblock
  \href{https://forum.mafiascum.net/viewtopic.php?f=5\&t=79490}{{Newbscum
  Greeting Tell {\ifmmode\bullet\else\textbullet\fi} Mafiascum.net}}.
\newblock [Online; accessed 26. Apr. 2019].
\newblock
  \href{https://forum.mafiascum.net/viewtopic.php?f=5\&t=79490}{https://forum.mafiascum.net/viewtopic.php?f=5\&t=79490}.

\bibitem[{Mehrabian(1971)}]{Mehrabian:1971}
Albert Mehrabian. 1971.
\newblock Nonverbal betrayal of feeling.
\newblock {\em Journal of Experimental Research in Personality\/} 5(1):64--73.

\bibitem[{Migdal(2010)}]{Migdal:2010}
Piotr Migdal. 2010.
\newblock \href{http://arxiv.org/abs/1009.1031v3}{A mathematical model of the
  mafia game}
  \href{http://arxiv.org/abs/1009.1031v3}{http://arxiv.org/abs/1009.1031v3}.

\bibitem[{Mihalcea and Strapparava(2009)}]{Mihalcea:2009}
Rada Mihalcea and Carlo Strapparava. 2009.
\newblock The lie detector: Explorations in the automatic recognition of
  deceptive language.
\newblock In {\em Proceedings of the ACL-IJCNLP 2009 Conference Short
  Papers\/}. Association for Computational Linguistics, pages 309--312.

\bibitem[{Mihaylov and Nakov(2016)}]{Mihaylov:2016}
Todor Mihaylov and Preslav Nakov. 2016.
\newblock Hunting for troll comments in news community forums.
\newblock In {\em Proceedings of the 54th Annual Meeting of the Association for
  Computational Linguistics\/}. pages 399--405.

\bibitem[{Pak and Zhou(2012)}]{Pak:2012}
Jinie Pak and Lina Zhou. 2012.
\newblock A social network based analysis of deceptive communication in online
  chat.
\newblock In {\em E-Life: 10th Workshop on E-Business\/}. Springer, Berlin,
  Heidelberg, pages 55--65.

\bibitem[{P{\'e}rez-Rosas et~al.(2015)P{\'e}rez-Rosas, Abouelenien, Mihalcea,
  and Burzo}]{perez2015deception}
Ver{\'o}nica P{\'e}rez-Rosas, Mohamed Abouelenien, Rada Mihalcea, and Mihai
  Burzo. 2015.
\newblock Deception detection using real-life trial data.
\newblock In {\em Proceedings of the 2015 ACM on International Conference on
  Multimodal Interaction\/}. ACM, pages 59--66.

\bibitem[{P{\'e}rez-Rosas and Mihalcea(2015)}]{perez2015experiments}
Ver{\'o}nica P{\'e}rez-Rosas and Rada Mihalcea. 2015.
\newblock Experiments in open domain deception detection.
\newblock In {\em Proceedings of the 2015 Conference on Empirical Methods in
  Natural Language Processing\/}. pages 1120--1125.

\bibitem[{Rosenthal et~al.(2017)Rosenthal, Farra, and
  Nakov}]{rosenthal2017semeval}
Sara Rosenthal, Noura Farra, and Preslav Nakov. 2017.
\newblock Semeval-2017 task 4: Sentiment analysis in twitter.
\newblock In {\em Proceedings of the 11th International Workshop on Semantic
  Evaluation (SemEval-2017)\/}. pages 502--518.

\bibitem[{Toomai(2014)}]{toomai}
Toomai. 2014.
\newblock \href{https://forum.mafiascum.net/viewtopic.php?f=5\&t=39739}{{The
  Newbie Matrix6 stats thread (complete) {\ifmmode\bullet\else\textbullet\fi}
  Mafiascum.net}}.
\newblock [Online; accessed 5. Feb. 2019].
\newblock
  \href{https://forum.mafiascum.net/viewtopic.php?f=5\&t=39739}{https://forum.mafiascum.net/viewtopic.php?f=5\&t=39739}.

\bibitem[{Wang(2017)}]{wang2017liar}
William~Yang Wang. 2017.
\newblock " liar, liar pants on fire": A new benchmark dataset for fake news
  detection.
\newblock {\em arXiv preprint arXiv:1705.00648\/} .

\bibitem[{Zhou et~al.(2004)Zhou, Burgoon, Nunamaker, and Twitchell}]{Zhou:2004}
Lina Zhou, Judee~K. Burgoon, Jay~F. Nunamaker, and Doug Twitchell. 2004.
\newblock Automating linguistics-based cues for detecting deception in
  text-based asynchronous computer-mediated communications.
\newblock {\em Group Decision and Negotiation\/} 13(1):81--106.

\bibitem[{Zhou and Sung(2008)}]{Zhou:2008}
Lina Zhou and {Yu-wei} Sung. 2008.
\newblock Cues to deception in online chinese groups.
\newblock In {\em Proceedings of the 41st Hawaii International Conference on
  System Sciences\/}. pages 146--46.

\end{thebibliography}
\bibliographystyle{emnlp_natbib}

\newpage

\end{document}

\subsection{Appendix A: Raw Performance
Results}\label{appendix-a-raw-performance-results}

\scalebox{0.83}{
\begin{tabular}{llllrr}
\toprule
      &       &      &               &  ROC AUC &     PR \\
replacements & wc & N & name &          &        \\
\midrule
True  & 50    & 9676 & Hand-picked &    0.565 &  0.277 \\
      &       &      & HP + FT wiki &    0.588 &  0.291 \\
      & 1000  & 7084 & Hand-picked &    0.581 &  0.289 \\
      &       &      & HP + FT wiki &    0.611 &  0.312 \\
      & 2000  & 5228 & Hand-picked &    0.591 &  0.297 \\
      &       &      & HP + FT wiki &    0.630 &  0.333 \\
      & 3000  & 3964 & Hand-picked &    0.590 &  0.307 \\
      &       &      & HP + FT wiki &    0.630 &  0.339 \\
      & 4000  & 3062 & Hand-picked &    0.589 &  0.315 \\
      &       &      & HP + FT wiki &    0.625 &  0.339 \\
      & 5000  & 2401 & Hand-picked &    0.594 &  0.319 \\
      &       &      & HP + FT wiki &    0.634 &  0.356 \\
      &       &      & FastText wiki &    0.627 &  0.342 \\
      &       &      & GloVe wiki200d &    0.624 &  0.332 \\
      &       &      & GloVe wiki &    0.624 &  0.331 \\
      &       &      & GloVe twitter &    0.624 &  0.335 \\
      & 6000  & 1935 & Hand-picked &    0.581 &  0.310 \\
      &       &      & HP + FT wiki &    0.613 &  0.343 \\
      & 7000  & 1577 & Hand-picked &    0.583 &  0.309 \\
      &       &      & HP + FT wiki &    0.614 &  0.336 \\
      & 8000  & 1266 & Hand-picked &    0.555 &  0.283 \\
      &       &      & HP + FT wiki &    0.594 &  0.322 \\
      & 9000  & 1039 & Hand-picked &    0.561 &  0.299 \\
      &       &      & HP + FT wiki &    0.580 &  0.322 \\
      & 10000 & 863  & Hand-picked &    0.579 &  0.307 \\
      &       &      & HP + FT wiki &    0.596 &  0.332 \\
False & 50    & 6437 & Hand-picked &    0.567 &  0.271 \\
      &       &      & HP + FT wiki &    0.590 &  0.285 \\
      & 1000  & 5919 & Hand-picked &    0.574 &  0.290 \\
      &       &      & HP + FT wiki &    0.603 &  0.312 \\
      & 2000  & 4728 & Hand-picked &    0.586 &  0.298 \\
      &       &      & HP + FT wiki &    0.624 &  0.331 \\
      & 3000  & 3724 & Hand-picked &    0.589 &  0.306 \\
      &       &      & HP + FT wiki &    0.629 &  0.339 \\
      & 4000  & 2943 & Hand-picked &    0.586 &  0.315 \\
      &       &      & HP + FT wiki &    0.621 &  0.340 \\
      & 5000  & 2320 & Hand-picked &    0.595 &  0.319 \\
      &       &      & HP + FT wiki &    0.634 &  0.351 \\
      &       &      & FastText wiki &    0.628 &  0.349 \\
      &       &      & GloVe wiki200d &    0.624 &  0.325 \\
      &       &      & GloVe wiki &    0.623 &  0.329 \\
      &       &      & GloVe twitter &    0.621 &  0.330 \\
      & 6000  & 1887 & Hand-picked &    0.583 &  0.313 \\
      &       &      & HP + FT wiki &    0.614 &  0.343 \\
      & 7000  & 1541 & Hand-picked &    0.580 &  0.309 \\
      &       &      & HP + FT wiki &    0.610 &  0.341 \\
      & 8000  & 1240 & Hand-picked &    0.550 &  0.278 \\
      &       &      & HP + FT wiki &    0.586 &  0.316 \\
      & 9000  & 1025 & Hand-picked &    0.549 &  0.292 \\
      &       &      & HP + FT wiki &    0.570 &  0.315 \\
      & 10000 & 857  & Hand-picked &    0.576 &  0.300 \\
      &       &      & HP + FT wiki &    0.591 &  0.323 \\
\bottomrule
\end{tabular}
}